\documentclass[letterpaper, 10 pt, journal, twoside]{IEEEtran}

\IEEEoverridecommandlockouts                              

\usepackage{times}

\usepackage[numbers]{natbib}
\usepackage{multicol}
\usepackage[bookmarks=true]{hyperref}

\usepackage{tikz}
\usepackage{comment}
\usepackage{amsmath,amssymb} 
\usepackage{color}
\usepackage{wrapfig}

\usepackage{url}
\usepackage{graphicx}
\usepackage{booktabs}
\usepackage{multirow}
\usepackage{vcell}

\usepackage{array}
\usepackage{makecell}
\usepackage{rotating}
\usepackage{arydshln}
\usepackage{soul} 
\usepackage{tikz}

\begin{document}

\newcommand{\ch}{\textcolor{black}}
\newcommand\copyrighttext{%
  \footnotesize \textcopyright 2023 IEEE. Personal use of this material is permitted.
  Permission from IEEE must be obtained for all other uses, in any current or future
  media, including reprinting/republishing this material for advertising or promotional
  purposes, creating new collective works, for resale or redistribution to servers or
  lists, or reuse of any copyrighted component of this work in other works.}
\newcommand\copyrightnotice{%
\begin{tikzpicture}[remember picture,overlay]
\node[anchor=south,yshift=2pt] at (current page.south) {\fbox{\parbox{\dimexpr\textwidth-\fboxsep-\fboxrule\relax}{\copyrighttext}}};
\end{tikzpicture}%
}

\title{\LARGE \bf
KINet: Unsupervised Forward Models \\ for Robotic Pushing Manipulation}

\author{Alireza Rezazadeh, Changhyun Choi
\thanks{Authors are with the Department of Electrical and Computer Engineering, University of Minnesota, Minneapolis, MN 55414 USA ({\tt\small rezaz003@umn.edu}, {\tt\small cchoi@umn.edu}). (Corresponding author: Alireza Rezazadeh.)}%
}

\author{Alireza Rezazadeh and Changhyun Choi%
\thanks{Manuscript received: April, 27, 2023; Revised July, 12, 2023; Accepted August, 2, 2023.}
\thanks{This paper was recommended for publication by Editor Hong Liu upon evaluation of the Associate Editor and Reviewers' comments.} 
\thanks{Authors are with the Department of Electrical and Computer Engineering, University of Minnesota, Minneapolis, MN 55414 USA ({\tt\small rezaz003@umn.edu}, {\tt\small cchoi@umn.edu}).}%
}

\markboth{IEEE Robotics and Automation Letters. Preprint Version. Accepted August, 2023}
{Rezazadeh \MakeLowercase{\textit{et al.}}: Unsupervised Forward Models for Robotic Pushing Manipulation} 

\maketitle
\copyrightnotice

\begin{abstract}
Object-centric representation is an essential abstraction for forward prediction. Most existing forward models learn this representation through extensive supervision (e.g., object class and bounding box) although such ground-truth information is not readily accessible in reality. To address this, we introduce KINet (Keypoint Interaction Network)---an end-to-end unsupervised framework to reason about object interactions based on a keypoint representation. Using visual observations, our model learns to associate objects with keypoint coordinates and discovers a graph representation of the system as a set of keypoint embeddings and their relations. It then learns an action-conditioned forward model using contrastive estimation to predict future keypoint states. By learning to perform physical reasoning in the keypoint space, our model automatically generalizes to scenarios with a different number of objects, novel backgrounds, and unseen object geometries. Experiments demonstrate the effectiveness of our model in accurately performing forward prediction and learning plannable object-centric representations for downstream robotic pushing manipulation tasks.
\end{abstract}


\IEEEpeerreviewmaketitle

\section{Introduction}
\IEEEPARstart{D}{iscovering} a structured representation of the world allows humans to perform a wide repertoire of motor tasks such as interacting with objects. The core of this process is learning to predict the response of the environment to applying an action \citep{miall1996forward}. The internal models, often referred to as the forward models, come up with an estimation of the future states of the world given its current state and the action. By cascading the predictions of a forward model, it is possible to plan a sequence of actions that would bring the world from an initial state to a desired goal state.

\begin{figure}[ht]
\begin{center}
    \includegraphics[width=0.9 \linewidth]{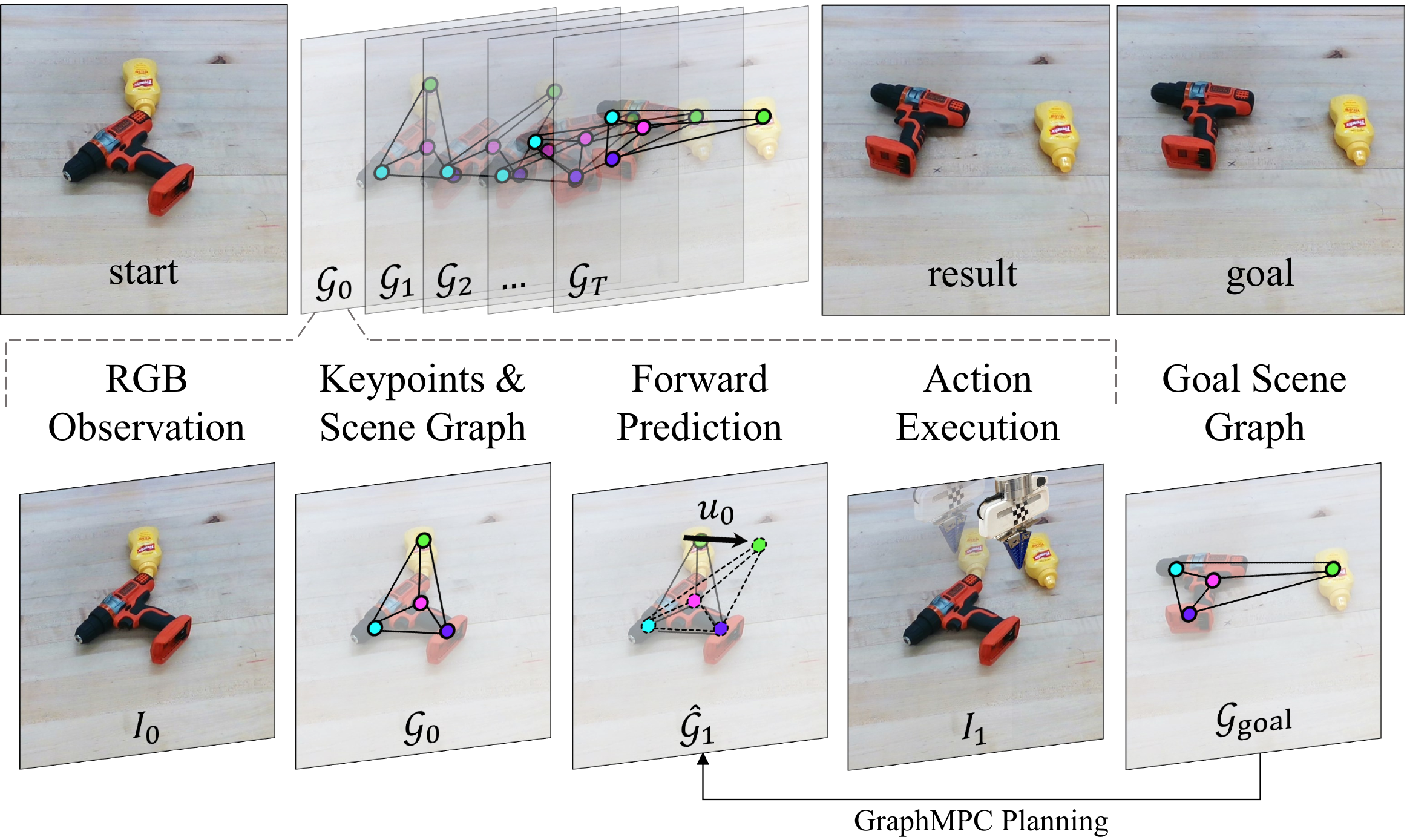}
    \vspace{-5pt}
    \caption{Existing forward modeling methods rely on the ground-truth object states, limiting their application to the real setting where such data is hard to obtain. Our framework (KINet) learns unsupervised forward models from raw RGB images. At each timestep, using the image observation ($I_t$), keypoints are extracted to construct the scene graph ($\mathcal{G}_t$). The graph-based forward model estimates the future keypoints ($\hat{\mathcal{G}}_{t+1}$) conditioned on the action $u_t$. Using this forward model, we perform model-based planning and find a sequence of actions to achieve a goal configuration based on similarity to the goal graph.}
    \vspace{-1mm}
    \label{fig:intuition}
    \end{center}
\end{figure}

Recently, deep learning architectures have been proposed to perform forward prediction using an object-centric representation of the system \citep{chen2021grounding, Li2020, qi2020learning, ye2020object}. This representation is learned from the visual observation by factorizing the scene into the underlying object instances using ground-truth object states (e.g., object class, position, and bounding box). We identified two major limitations in the existing work:
First, they either assume access to the ground-truth object states \citep{battaglia2016interaction, Li2020} or predict them using pre-trained object detection or instance segmentation models \citep{qi2020learning, ye2020object}. However, obtaining the ground truth object states is not always feasible in practice, especially in real-world settings. Relying on pre-trained object detection and segmentation models further makes the forward models fragile as the perception models do not work on novel objects, significantly limiting their generalization capability. Second, factorizing the scene into a fixed number of object instances limits the generalization of the model models to scenarios with a different number of objects. 

In this paper, we address both of these limitations by proposing to learn forward models using a keypoint representation. Keypoints represent a set of salient locations of moving entities. Our model KINet (Keypoint Interaction Network) learns an unsupervised forward model in three steps (Fig \ref{fig:intuition}): (1) A keypoint extractor factorizes the scene into keypoints with no supervision other than raw visual observations. (2) A graph representation of the scene is learned, where each node corresponds to a keypoint and edges are keypoints relations. Node features carry implicit object-centric features as well as explicit keypoint state information. (3) With probabilistic message passing, our model learns an action-conditional forward model to predict the future location of keypoints and reconstruct the future appearances of the scene. We evaluate KINet's forward prediction accuracy and demonstrate that, by learning forward prediction in a keypoint coordinate, our model effectively re-purposes this knowledge and generalizes it to complex unseen circumstances.

Our key contributions are: (1) We introduce KINet, a graph-based and end-to-end method for learning unsupervised action-conditional forward models from visual observations (2) We introduce probabilistic message-passing operation for efficient aggregation of relevant information in the graph. (3) We introduce GraphMPC for accurate action planning using graph similarity. (4) We demonstrate learning forward models with keypoints enables generalization to complex unseen scenarios.

\section{Related Work}
\noindent\textbf{Unsupervised keypoint extraction.} Keypoints have been widely used in pose tracking \citep{jafarian2018monet, zhang2018unsupervised} and video prediction \citep{manuelli2020keypoints, minderer2019unsupervised, xue2016visual, zhang2018unsupervised}. Recent work explored keypoints for reinforcement learning in a keypoint space \citep{chen2021unsupervised, jakab2018unsupervised, kulkarni2019unsupervised}.\\
\textbf{Forward models.} 
The most fundamentally relevant work to ours is Interaction Networks (IN) \citep{battaglia2016interaction, sanchez2018graph} and follow-up work using graph networks for forward simulation \citep{kipf2018neural, li2018propagation, mrowca2018flexible}. 
These methods rely on the ground-truth state of objects to build graphs where each node represents an object. 
Several approaches extended IN by combining explicit states with implicit visual features \citep{qi2020learning, watters2017visual, ye2020object}. 
However, two main concerns remain unaddressed. 
First, object features in the graph are often obtained from extensive ground-truth information such as position, bounding box, and mask \citep{kipf2018neural, Li2018, qi2020learning, ye2020object}. 
Second, these approaches lack generalization to varying number of objects as they are formulated on fixed-size graphs where each node has to correspond to one of the objects. To address these limitations, our framework extracts unsupervised keypoints to build a scene graph and performs forward prediction in the keypoint space. This allows performing forward modeling without any supervision other than RGB images. Also, factorizing the scene to keypoints instead of objects allows for automatic generalization to varying number of objects.
\\
\textbf{Action-conditional forward models.}  
\citet{battaglia2016interaction} augments the action to the node embeddings. 
\citet{ye2020object} included the action as an additional node in a fully connected graph while other nodes represent objects. 
For probabilistic forward models, \citet{henaff2019model} suggests using a latent variable dropout to condition the model on action \citep{gal2016dropout}. 
\citet{yan2020learning} showed the effectiveness of contrastive estimation \citep{oord2018representation} to learn actionable representations. 
\citet{minderer2019unsupervised} uses keypoints for video prediction from a history of frames. However, their dynamics model is not conditioned on action and cannot be used for action planning. 
In our work, we ensure conditioning on the action by incorporating the action vector in the graph message-passing operation. In addition, we extend the contrastive estimation method for graph embeddings to further enhance the learned representations.
\\
\textbf{Unsupervised Forward Models.} 
\citet{kipf2019contrastive} uses a contrastive loss to learn object-centric abstractions in multibody environments of fixed objects with minimal visual features such as 2D shapes. In our work, we randomize the properties of the objects and examine challenging realistic objects.
\citet{kossen2019structured} uses images to infer a set of object states (e.g., position and velocity) to predict the future state of each object using graph networks. Although learning unsupervised states, this method is formulated on a fixed number of objects and only tested on environments with simple 2D geometries. 
\citet{li2020causal} infers the causal graph representation and makes future predictions on a fixed dynamic system in simulation using a pretrained keypoint extractor from topview images. In our work, we experiment with other camera angles and real-world 3D objects with various properties such as geometry and texture.

\section{Keypoint Interaction Networks (KINet)}

\begin{figure*}[t]
\begin{center}
\includegraphics[width=1.00\linewidth]{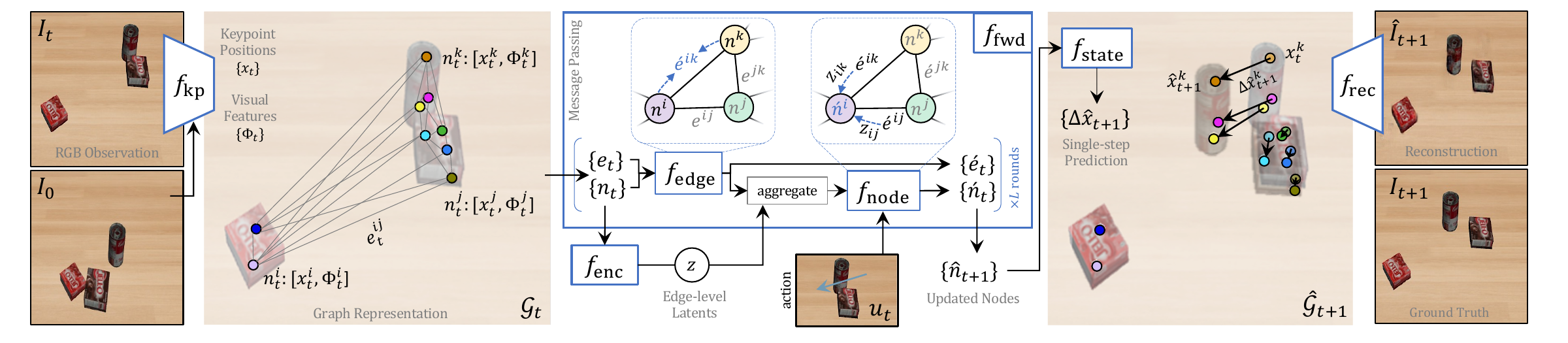}
\vspace{-10 mm}
\end{center}
\caption{\ch{Keypoint Interaction Network (KINet): A keypoint detector $f_\mathrm{kp}$ utilizes the current RGB image $I_t$ and the initial frame $I_0$ to identify a set of unsupervised keypoints $\{x_t\}$ and their corresponding visual features $\{\Phi_t\}$. A scene graph $\mathcal{G}t$ is constructed, where the keypoints are represented as nodes. The forward model $f_\mathrm{forw}$ processes the graph and action $u_t$ through $L$ rounds of probabilistic message-passing operations using the edge function $f_\mathrm{edge}$ and the node function $f_\mathrm{node}$, with the edge probabilities generated by an encoder $f_\mathrm{enc}$. Subsequently, a state decoder estimates the future positions of the keypoints $\hat{x}_{t+1}$. Finally, the appearance of the future scene is reconstructed by $f_\mathrm{rec}$.}}
\label{fig:network}
\end{figure*}

We assume access to RGB image, action vector, and the image after applying the action: $\mathcal{D} = \{(I_t, u_t, I_{t+1})\}$. Our goal is to learn a forward model that predicts the future states of the objects. We describe our approach in two main steps (see Fig.~\ref{fig:network}): learning to encode visual observations into keypoints and learning an action-conditioned forward model in the keypoint space.

\subsection{Unsupervised Keypoint Detection}
The keypoint detector ($f_{\mathrm{kp}}$, Fig.~\ref{fig:network}) is a mapping from RGB visual observations of the scene to a lower-dimensional set of $K$ keypoint coordinates $\{x_{t}^k\}_{k=1 \dotsc K} = f_{ \mathrm{kp}}(I_t)$. The keypoint coordinates are learned by capturing the spatial appearance of the objects in an unsupervised manner. 
Specifically, the keypoint detector receives a pair of initial and current images $(I_0, I_t)$ and uses a convolutional encoder to compute a $K$-dimensional feature map for each image $\Phi(I_0), \Phi(I_t) \in \mathbb{R}^{H^\prime\times W^\prime\times K}$. The expected number of keypoints is set by the dimension $K$. Next, the feature maps are marginalized into a 2D keypoint coordinate $\{x_{0}^k, x_{t}^k\}_{k=1 \dotsc K}\in \mathbb{R}^{2}$. We use a convolutional image reconstruction model ($f_{ \mathrm{rec}}$, Fig.~\ref{fig:network}) with skip connections to inpaint the current image frame using the initial image and the predicted keypoint coordinates $\hat{I}_t = f_{ \mathrm{rec}}(I_0,\{x_{0}^k, x_{t}^k\}_{k=1 \dotsc K})$. 
With this formulation, $f_{\mathrm{kp}}$ and $f_{\mathrm{rec}}$ create a bottleneck to encode the visual observation in a temporally consistent lower-dimensional set of keypoints \citep{kulkarni2019unsupervised}. 

\subsection{Graph Representation of Scene}
After factorizing the scene into $K$ keypoints, we build a graph $\mathcal{G}_t = (\mathcal{V}_t, \mathcal{E}_t, \mathbf{Z})$ (undirected, no self-loop) where keypoints and their pairwise relations are the graph nodes and edges. Keypoint positional and visual information are encoded into embedding of nodes $\{\mathbf{n}_t^k\}_{k=1 \dotsc K} \in \mathcal{V}_t$ and edges $\{\mathbf{e}_t^{ij}\} \in \mathcal{E}_t$. We define the adjacency matrix to specify the graph connectivity as $\mathbf{Z} \in \mathbb{R}^{K\times K}$ where $z_{ij} \in [0,1]$ is the probability of the edge $\mathbf{e}^{ij} \in \mathcal{E}$. \ch{At timestep $t$, node embeddings encode keypoint positional and visual features $\mathbf{n}_t^k = \big[ \Phi_t^k, x_t^k\big]$. Edge embeddings contain relative positional information of each node pair $\mathbf{e}_t^{ij} = \big[x_t^{i}-x_t^{j}, \|x_t^{i}-x_t^{j}\|_2^2 \big]$.}

Existing graph-based forward modeling methods construct the scene graph such that each node represents an object. The node embeddings are positional and visual features extracted using ground-truth object state information \citep{kipf2018neural, qi2020learning, ye2020object}. Our framework, however, does not rely on any ground-truth information supervision and only uses RGB images.

\subsection{\ch{Probabilistic Graph-based Forward Model}}
\ch{We extend the existing graph-based approaches   \cite{battaglia2016interaction, sanchez2018graph} and propose probabilistic graph-based forward models. The core of our forward model is probabilistic message-passing using edge-level latent variables $z_{ij}\in \mathbb{R}^{d}$ that represent the edge probabilities.} A posterior network $p_{\theta}$ infers the elements of the adjacency matrix given the scene graph representation. In particular, we model the posterior as $p_{\theta}(z_{ij} | \mathcal{G}_t) = \sigma(f_{\mathrm{enc}}([\mathbf{n}_t^i, \mathbf{n}_t^j]))$ where $\sigma(.)$ is the sigmoid function. 

\ch{The forward model $\hat{\mathcal{G}}_{t+1} = f_{ \mathrm{fwd}}(\mathcal{G}_t, u_t, \mathbf{Z})$ predicts the graph representation at the next timestep by taking as input the current graph representation and the action vector ($f_{\mathrm{fwd}}$, Fig.~\ref{fig:network}) and performing $L$ rounds of probabilistic message-passing. Fig.~\ref{fig:network}-$f_\mathrm{fwd}$ illustrates the probabilistic message-passing operation. A single round of message-passing operation in the forward model can be described as, 
\vspace{-3pt}\begin{align*}
    &\text{Edge update}: {\mathbf{e'}}^{ij} \leftarrow f_{\mathrm{edge}}({\mathbf{n}}^i, {\mathbf{n}}^j, {\mathbf{e}}^{ij}) \\
    &\text{Aggregation}: \bar{\mathbf{e}}^{k} \leftarrow \Sigma_{i\in N(k)}{z}_{ik}{\mathbf{e'}}^{ik} \\
    &\text{Node update}: {\mathbf{n'}}^{k} \leftarrow f_{\mathrm{node}}({\mathbf{n}}^k, \bar{\mathbf{e}}^{k}, u_t)
\end{align*}
where the edge-specific function \(f_ \mathrm{edge}\) first updates edge embeddings, then the node-specific function \(f_ \mathrm{node}\) updates each node embedding ${\mathbf{n'}}^{k}$ by aggregating its neighboring nodes $N(k)$ information. Specifically, the updated edges are aggregated using the edge probabilities from the inferred adjacency matrix ${\mathbf{Z}}$ as weights. The action $u_t$ is also an input to the aggregation step. After performing the $L$ rounds of message-passing, the resulting updated node embeddings estimate the graph's nodes at the next timestep $\{\hat{\mathbf{n}}_{t+1}\}$.}

Recent models for forward prediction rely on fully connected graphs for message passing \citep{Li2018, qi2020learning, ye2020object}. Our model, however, learns to probabilistically sample the neighbor information. Intuitively, this adaptive sampling allows the network to efficiently aggregate the most relevant neighboring information. This is specifically essential in our model as keypoints could provide redundant information if they are in close proximity.

\subsection{Forward Prediction}

The state decoder ($f_{\mathrm{state}}$, Fig.~\ref{fig:network}) transforms the predicted node embeddings of the updated graph $\hat{\mathcal{G}}_{t+1}$ to a first-order difference $\{\Delta \hat{x}_{t+1}^k\}=f_{\mathrm{state}}(\{\mathbf{\hat{n}}_{t+1}^{k}\})$, which is integrated once to predict the position of the keypoints in the next timestep $\{\hat{x}_{t+1}^k\} = \{x_{t}^k\} + \{\Delta \hat{x}_{t+1}^k\}$. To reconstruct the image at the next timestep, we borrow the reconstruction model \(f_\mathrm{rec}\) from the keypoint detector $\hat{I}_{t+1} = f_{ \mathrm{rec}}(I_0,\{x_{0}^k, \hat{x}_{t+1}^k\})$.

\subsection{Learning KINet}
\noindent\textbf{Reconstruction loss.}
The keypoint detector is trained to reconstruct the image at each timestep $\mathcal{L}_{\mathrm{rec}} = \| \hat{I}_t - I_t \|_2^2$. As suggested by \cite{minderer2019unsupervised}, errors from the keypoint detector were not backpropagated to other modules of the model. This ensures that the model does not conflate errors from keypoint extraction and forward prediction functions.\\
\textbf{Forward loss.}
The model is optimized to predict the next state of the keypoints. A forward loss penalizes the distance between the estimated future keypoint locations using first-order state decoder and the keypoint extractor predictions: $\mathcal{L}_{\mathrm{fwd}} = \sum_K \| \hat{x}_{t+1}^k - f_{ \mathrm{kp}}(I_{t+1})^k \| _2^2$. \\
\textbf{Inference loss.}
Our model is trained to minimize the KL-divergence between the posterior and prior: $\mathcal{L}_{\mathrm{infer}} = D_{\mathrm{KL}}\big(p_{\phi}(\mathbf{Z}|\mathcal{G}) \big\| p(\mathbf{Z})\big)$. We use independent Gaussian prior $p(\mathbf{Z}) = \prod_i\mathcal{N}(\mathbf{z}_i)$ and reparameterization for training \citep{kingma2013auto}.\\
\textbf{Contrastive loss.}
We use the contrastive estimation method to further enhance the learned graph representations. We add a contrastive loss \citep{oord2018representation, yan2020learning} and reframe it for graph embeddings as
$
     \mathcal{L}_{\mathrm{ctr}} = -\mathbb{E}_{\mathcal{D}}[\mathrm{log}({\mathcal{S}(\hat{\mathcal{G}}_{t+1}, \mathcal{G}^{+}_{t+1})}/{\sum\mathcal{S}(\hat{\mathcal{G}}_{t+1}, \mathcal{G}^{-}_{t+1})})]
$ such that the predicted graph representations $\hat{\mathcal{G}}_{t+1}$ are maximally similar to their corresponding positive sample pair ${\mathcal{G}}^{+}_{t+1} := {\mathcal{G}}_{t+1}$ and maximally distant from the negative sample pairs ${\mathcal{G}}^{-}_{t+1}:=G_{\tau} \ \ \forall \tau \neq t+1$. We use a simple node embedding similarity as the graph matching algorithm $\mathcal{S}(\mathcal{G}_1, \mathcal{G}_2) = \sum_K{\{n_1^k\}.\{n_2^k\}}$. The motivation behind adding a contrastive loss is aligning the graph representation of similar object configurations while pushing apart those of dissimilar configurations in the embedding space to enhance the learned graphs.
Finally, the combined loss is: $\mathcal{L} = \lambda_{\mathrm{rec}} \ \mathcal{L}_{\mathrm{rec}} + \lambda_{\mathrm{fwd}} \ \mathcal{L}_{\mathrm{fwd}} + \lambda_{\mathrm{infer}} \ \mathcal{L}_{\mathrm{infer}} +\ \lambda_{\mathrm{ctr}} \ \mathcal{L}_{\mathrm{ctr}}.$
\begin{figure*}[ht]
\begin{center}
    \includegraphics[width=1.00\linewidth]{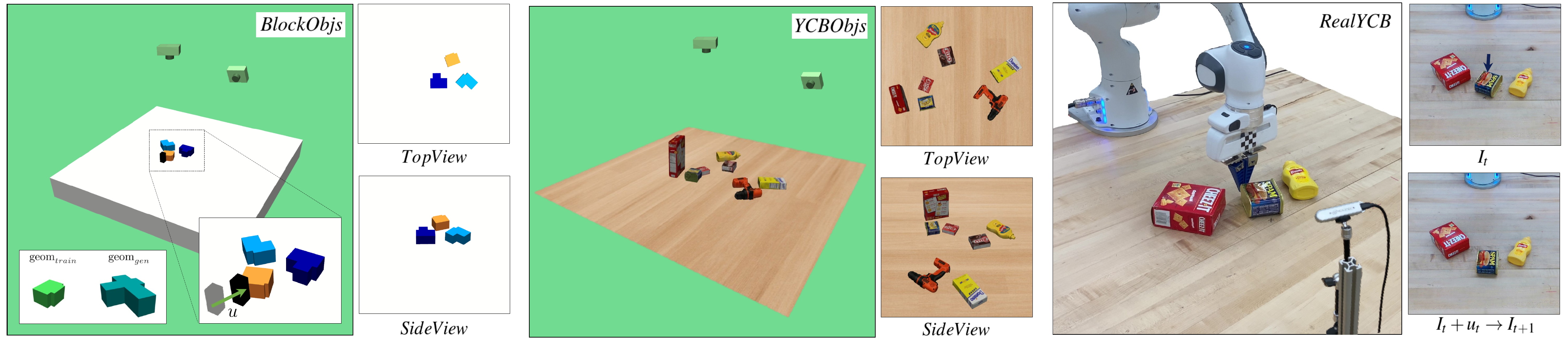}
    \caption{Experiment setups including two simulated environments with block objects and YCB objects, and a real robot environment with YCB objects. We trained our model in simulation on 3 objects and then tested for their generalization to unseen geometries and an unseen number of objects.}
    \label{fig:setup}
    \end{center}
\end{figure*}

\subsection{GraphMPC Planning with KINet}
\label{section:mpc}

A learned KINet model can be used to perform model-based planning. We extend the Model Predictive Control (MPC) method \citep{garcia1989model} and propose the GraphMPC algorithm based on graph embeddings. \ch{The graph representation of the next timestep is estimated for multiple sampled actions using $f_\mathrm{fwd}$}. The optimal action is selected such that it produces the most similar graph representation to the goal graph representation $\mathcal{G}^{\mathrm{goal}}$. We describe our GraphMPC algorithm with a horizon of $T$ as: $ {u_t^* = {\arg\max}\{\mathcal{S}\big(\mathcal{G}^{\mathrm{goal}}, f_{ \mathrm{fwd}}(\mathcal{G}^t, \{u_{t:T}\})\big)\}; \ t \in [0, T] }$. Unlike performing conventional MPC only with respect to positional states, GraphMPC allows for accurately bringing the objects to a goal configuration both explicitly (i.e., position) and implicitly (i.e., pose, orientation, and visual appearance).

\section{Experimental Setups}
Our experiments address the following: (1) How accurate is our forward model? (2) Can the model be used for action planning? (3) Does the model generalize to unseen scenarios?

\subsection{Dataset} We apply our approach to learn a forward model for multi-object manipulation tasks. The task involves rearranging the objects to achieve a goal configuration using pushing actions. We use MuJoCo 2.0 \citep{todorov2012mujoco} to generate two simulated environments. We also test on a real robot environment. Fig.~\ref{fig:setup} demonstrates these three experiment setups.

\noindent\textit{\textbf{BlockObjs}}: Each object is constructed with two cuboid geoms. We sample the geom dimensions from a continuous range denoted as $\mathrm{geom}_{train}$ for training and $\mathrm{geom}_{gen}$ for generalization. Unseen geometries are designed to have elongated shapes to create extreme out-of-distribution cases (see Fig.~\ref{fig:setup}).

\noindent\textit{\textbf{YCBObjs}}: A subset of YCB objects placed on a wooden table that includes objects of daily life with diverse properties such as shape, size, and texture \citep{calli2015ycb}  (see Fig.~\ref{fig:setup}). 

In both of the simulated environments, we generate 10K episodes of random pushing on multiple objects (1-5 objects) where a simplified robot end-effector applies randomized pushing for 60 timesteps per episode. We collect the 4-dimensional action vector (pushing start and end location) and RGB images before and after each action is applied. Images are obtained using an overhead (\textit{TopView}) and an angled camera (\textit{SideView}).

\noindent\textit{\textbf{RealYCB}}: We directly transfer the learned model from the \textit{YCBObjs} simulation to the real environment with YCB objects. The real robot setup includes a Franka Emika Panda robot with Festo soft fingers and an Intel RealSense camera on the side of the tabletop (see Fig.~\ref{fig:setup}).

\subsection{Baselines} We compare our approach with existing methods for learning object-centric forward models:

\noindent\textbf{Forward Model} \textit{(Forw)}: We train a convolutional encoder to extract visual features of the scene image ({\textit{Img}}) and learn a forward model in the feature space. \\
\textbf{Forward-Inverse Model}~\cite{agrawal2016learning} \textit{(ForwInv)}: We train a convolutional encoder to extract visual features of the scene image ({\textit{Img}}) and jointly learn forward and inverse models. \\
\textbf{Interaction Network}~\cite{battaglia2016interaction, sanchez2018graph} \textit{(IN)}: We build an Interaction Network based on the ground truth location of the objects available in simulation. Each object representation contains the ground-truth position and velocity of the objects (\textit{GT state}). Note that this approach is only applicable to scenarios where the number of objects in the scene is known and fixed. \\
\textbf{Visual Interaction Network}~\cite{watters2017visual, ye2020object} \textit{(VisIN)}: We train a convolutional encoder to extract visual features of fixed-size bounding boxes centered on ground-truth object locations (\textit{GT state + Img}). We use the extracted visual features as object representations in the Interaction Network. Note that this approach requires prior knowledge of the number of objects. \\
\textbf{Causal Discovery from Videos}~\cite{li2020causal} \textit{(V-CDN)}: A pretrained perception module extracts keypoints that are used in an inference module to predict a causal structure of the visual observation which is then used in a dynamics module to predict the future location of the keypoints. 

\noindent\ch{We also compare KINet with two variants: (1) \textit{KINet - determ}, which employs a fully connected graph instead of probabilistic edges, and (2) \textit{KINet - no ctr}, which excludes the contrastive loss.}
 
\subsection{Training and Evaluation Setting.} 

All models are trained on a subset of the simulated data (\textit{BlockObjs} and \textit{YCBObjs}) with 3 objects (8K episodes: 80\% training, 10\% validation, and 10\% testing sets). To evaluate generalization to a different number of objects, we use other subsets of data with 1, 2, 4, and 5 objects ($\sim400$ episodes for each case). 
We train our model separately on images obtained from the overhead camera (\textit{TopView}) and the angled camera (\textit{SideView}). 
We set the expected number of keypoints $K=6$ for \textit{BlockObjs}, $K=9$ for \textit{YCBObjs} and \textit{RealYCB}. 
\ch{For the message-passing operation, the number of rounds is set to $L=3$. Note that $f_\mathrm{enc}$, $f_ \mathrm{edge}$, and $f_ \mathrm{node}$ are MLPs with three 128-dim hidden layers.}

\begin{figure*}[ht]
  \begin{center}
    \includegraphics[width=1.00\linewidth]{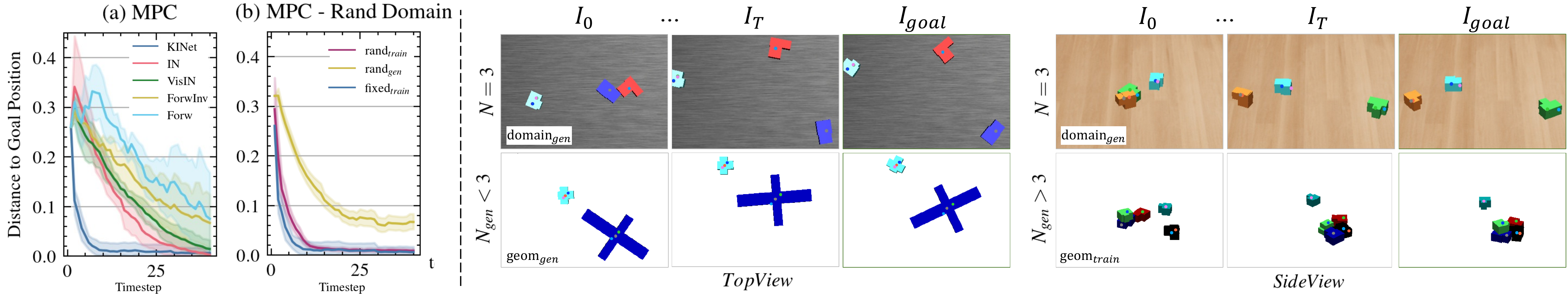}
  \end{center}
  \caption{MPC on \textit{BlockObjs}. Left panel: MPC result measured as the distance to goal configuration. (a) Comparison with baselines. (b) MPC for KINet trained on a fixed white background ($\mathrm{fixed}_{train}$), generalization to random backgrounds ($\mathrm{rand}_{gen}$), and trained on random backgrounds ($\mathrm{rand}_{train}$). Right panel: qualitative examples of generalization to unseen geometries, unseen number of objects, and unseen table textures.}
\label{fig:blocks-mpc} 
\end{figure*}
\section{Results}
This section is organized to thoroughly evaluate our model in simulated and real environments.  


\subsection{Does the model accurately learn a forward model?} 
First, we evaluate the forward prediction accuracy. Our model factorizes the observation into a set of keypoint representations and accurately estimates the future appearance of the scene conditioned on external action.

\begin{table}[h]
\caption{Forward Prediction performance on \textit{BlockObjs} measured as single-step mean predictions error {\tiny$\times 10^{-3}$}.}
\vspace{-3 mm}
\centering
\begin{tabular}{lcc|cc}
\toprule
Model        & \textit{SideView}                                  & \textit{TopView}                                   & Img   & GT state\\ \hline
IN           &  $0.109 \scriptscriptstyle\pm 0.01$         &$0.112 \scriptscriptstyle\pm 0.01$        & -           & \checkmark\\
VisIN        &  $0.121 \scriptscriptstyle\pm 0.03$         &$0.107 \scriptscriptstyle\pm 0.01$        & \checkmark  & \checkmark\\ \hdashline
Forw         &  $0.317 \scriptscriptstyle\pm 0.09$         &$0.309 \scriptscriptstyle\pm 0.12$        & \checkmark  & -\\
ForwInv      &  $0.266 \scriptscriptstyle\pm 0.02$         &$0.293 \scriptscriptstyle\pm 0.08$        & \checkmark  & -\\
\textbf{KINet} (ours) &  {$\mathbf{0.129} \scriptscriptstyle\pm 0.02$}         &\textbf{$\mathbf{0.122} \scriptscriptstyle\pm 0.01$}        & \checkmark  & -\\ \hdashline
KINet - determ &  $0.133 \scriptscriptstyle\pm 0.01$         &$0.127 \scriptscriptstyle\pm 0.03$        & \checkmark  & -\\ 
KINet - no ctr &  $0.169 \scriptscriptstyle\pm 0.05$         &$0.173 \scriptscriptstyle\pm 0.02$        & \checkmark  & -\\ 
\bottomrule    
\end{tabular}\label{table:forw}
\vspace{-2 mm}
\end{table}

Using \textit{BlockObjs} data, we quantify the effectiveness of our model in comparison with Forw, ForwInv, IN, and VisIN baselines (Table \ref{table:forw}). We separately train and examine each model on \textit{TopView} and \textit{SideView} images. The prediction error is computed as the average distance between the predicted and ground-truth positional states. 
VisIN performs the best among baseline models as it builds object representations with explicit ground-truth object positions and their visual features. Our model, on the other hand, achieves a comparable performance to VisIN while it does not rely on any supervision beyond the scene images. Forw and ForwInv baselines have similar supervision to ours but are significantly less accurate. This emphasizes the capability of our approach in learning a rich graph representation of the scene and an accurate forward model, while relaxing prevailing assumptions of the prior work on the structure of the environment and the availability of ground-truth state information.
KINet - determ \& no ctr will be discussed in Section~\ref{sec:results:ablation}.
\subsection{Can we use the model in control tasks?}

We design a robotic manipulation task of rearranging a set of objects to the desired goal state using MPC with pushing actions. For all models, we run 1K episodes with randomized object geometries and initial poses and a random goal configuration of objects. The planning horizon is set to $T=40$ timesteps in each episode. For our model, we perform GraphMPC based on graph embedding similarity as described in Section \ref{section:mpc}. For all baseline models, we perform MPC directly on the distance to the goal. Fig.~\ref{fig:blocks-mpc}a shows MPC results of \textit{BlockObjs} based on \textit{Top View} observations. Our approach is consistently more accurate and faster in reaching the goal configuration compared to the baseline models. 

\subsection{Does the model generalize to unseen circumstances?} 

One of our main motivations to learn a forward model in the keypoint space is to eliminate the dependency of model formulation on the number of objects in the scene which allows for generalization to an unseen number of objects, object geometries, background textures, etc.

\noindent\textit{\textbf{BlockObjs}}. We train KINet on 3 randomized blocks with ($\mathrm{geom}_{train}$) and test for generalization to an unseen number of objects (1, 2, 4, 5), unseen object geometries ($\mathrm{geom}_{gen}$). Fig.~\ref{fig:blocks-mpc} shows qualitative generalization results. We separately train and examine for generalization on \textit{TopView} and \textit{SideView} images. Since our model learns to perform forward modeling in the keypoint space, it reassigns the keypoints to unseen objects and makes forward predictions. Table \ref{table:gen} summarizes the generalization performance to unseen number of objects and unseen geometries. As expected, by increasing the number of objects the average distance to the goal position increases. Also, objects with out-of-distribution geometries ($\mathrm{geom}_{gen}$) have more distance to the goal position. 
KINet - determ will be discussed in Section~\ref{sec:results:ablation}.

\begin{table}[h]
\centering
\caption{Generalization results measured as the average distance to the goal position {\tiny$\times 10^{-3}$}.}  \vspace{-3 mm}
\begin{tabular}{c|cc|cc}
\toprule
  & \multicolumn{2}{c}{$\mathrm{geom}_{train}$}               & \multicolumn{2}{c}{$\mathrm{geom}_{gen}$}                 \\\cmidrule(lr){2-3}\cmidrule(lr){4-5}
N & KINet                       & KINet - determ               & KINet                       & KINet - determ               \\\hline
1 & $0.21 \scriptstyle\pm 0.04$ & $0.28 \scriptstyle\pm 0.01$ & $0.35 \scriptstyle\pm 0.04$ & $0.36 \scriptstyle\pm 0.08$ \\
2 & $0.21\scriptstyle\pm 0.03$  & $0.22 \scriptstyle\pm 0.05$ & $0.53 \scriptstyle\pm 0.06$ & $0.59 \scriptstyle\pm 0.07$ \\
3 & $0.19 \scriptstyle\pm 0.02$ & $0.19 \scriptstyle\pm 0.03$ & $0.20 \scriptstyle\pm 0.05$ & $0.31 \scriptstyle\pm 0.08$ \\
4 & $0.51 \scriptstyle\pm 0.02$ & $0.57 \scriptstyle\pm 0.12$ & $0.65 \scriptstyle\pm 0.07$ & $0.96 \scriptstyle\pm 0.09$ \\
5 & $0.89 \scriptstyle\pm 0.13$ & $1.05 \scriptstyle\pm 0.16$ & $1.64 \scriptstyle\pm 0.11$ & $2.17 \scriptstyle\pm 0.10$ \\
\bottomrule
\end{tabular}\label{table:gen}
\vspace{-3 mm}
\end{table}

\begin{figure*}[h]
    \begin{center}
    \includegraphics[width=\textwidth]{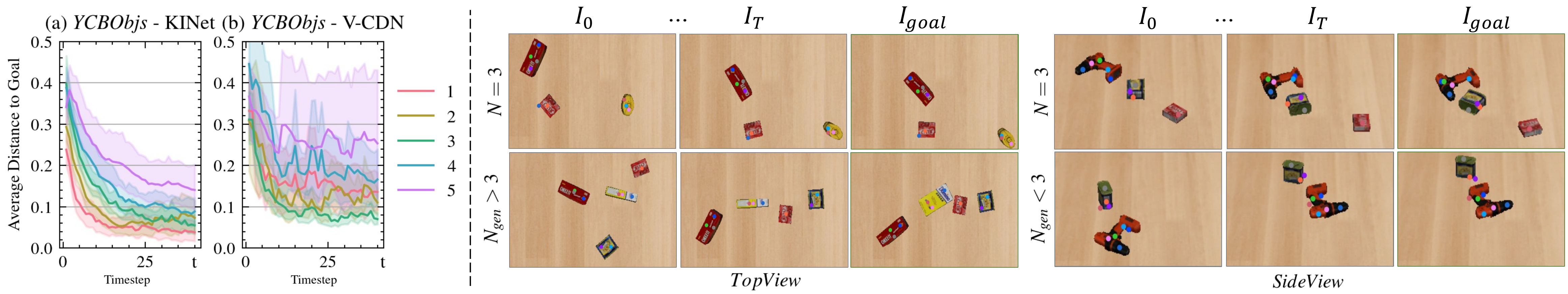}
    \vspace{-9 mm}
    \end{center}
    \caption{MPC for generalization to unseen number of objects on \textit{YCBObjs}. Left panel: baseline comparison. Right panel: results for \textit{Top} and \textit{SideView}.}
\label{fig:ycb-qual}\label{fig:ycb-mpc}
\vspace{-1 mm}
\end{figure*}

\begin{figure}[h]
    \begin{center}
    \includegraphics[width= 1.0\linewidth]{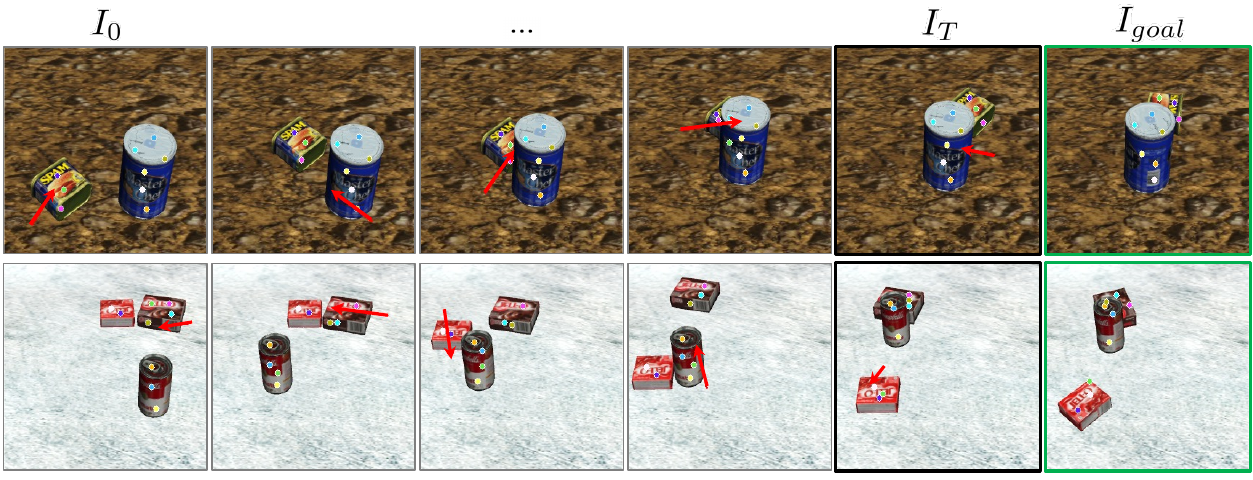}
\vspace{-9 mm}
    \end{center}
    \caption{KINet performance with challenging background textures and occlusion in the goal configurations.}
\label{fig:ycb-qual-hard}
\vspace{-1 mm}
\end{figure}

\begin{figure}[h]
\vspace{2 mm}
    \begin{center}
    \includegraphics[width= 1.\linewidth]{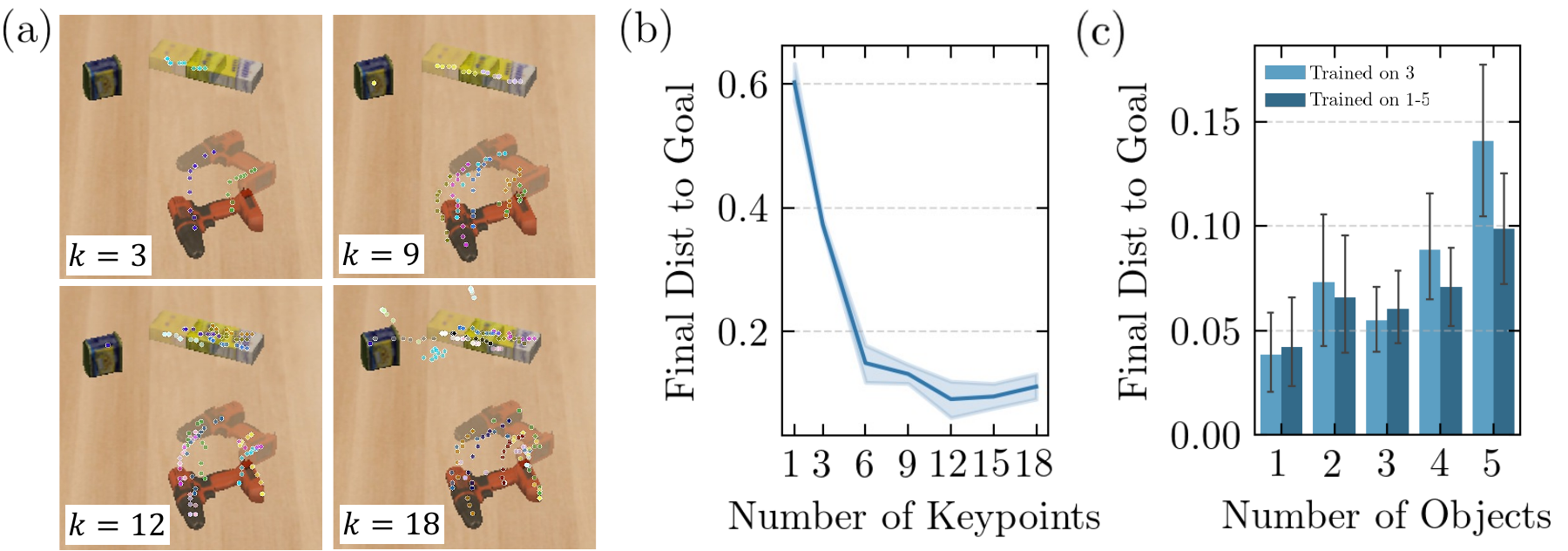}
    \vspace{-8 mm}
    \end{center}
    \caption{\ch{(a) Qualitative examples showcasing the predicted keypoints for varying $K$. (b) Distance to the goal configuration after 30 steps (T=30) for KINets trained with varying numbers of keypoints. (c) Distance to goal after 30 steps (T=30) for KINet trained on 3 objects compared to 1-5 objects.}}
\label{fig:var-kps} 
\end{figure}

Further, we test the performance of the model on unseen background textures (i.e., the table texture). Since the keypoint extraction relies on visual features of salient objects, our model is able to perform the control tasks by ignoring the background and assigning keypoints to the moving objects. Fig.~\ref{fig:blocks-mpc}b compares the MPC results for the KINet trained on a fixed white background ($\mathrm{fixed}_{train}$), zero-shot generalization of KINet trained on the fixed background to randomized backgrounds ($\mathrm{rand}_{gen}$), and KINet trained directly on randomized backgrounds ($\mathrm{rand}_{train}$).
As expected, although the MPC converges, the final distance to the goal configuration is larger for $\mathrm{rand}_{gen}$. This final error is statistically at the same level of accuracy for $\mathrm{fixed}_{train}$ and $\mathrm{rand}_{train}$. Qualitative examples are included in Fig.~\ref{fig:blocks-mpc}.

\noindent\textit{\textbf{YCBObjs}}. We train our model on a random subset of 3 YCB objects and test for generalization to an unseen number of objects (1,2,4,5). As shown in Fig.~\ref{fig:ycb-qual}, our method generalizes well to an unseen number of objects and performs the control task accurately. 
Importantly, assigning multiple keypoints to each object allows our framework to implicitly capture the orientation of each object, as well as their position without any supervision on the object pose (e.g., compare the power drill pose in Fig.~\ref{fig:ycb-qual}). Additional qualitative results for two challenging scenarios are included in Fig.~\ref{fig:ycb-qual-hard}. In these two examples, the tabletop is highly textured which shows the robustness of the keypoint factorization in binding to the moving object and ignoring the background. Moreover, objects in the goal configuration in these two examples are partially occluded. This means that even if an object is heavily occluded, our framework should still be able to find a sequence of actions to recover from the occluded configuration by leveraging the keypoints assigned to the non-occluded objects. We believe this is a significant advantage, as it allows our framework to handle challenging scenarios and achieve accurate performance.

We compare the performance of our framework with V-CDN \citep{li2020causal} baseline which is also a keypoint-based model to learn the structure of physical systems and perform future predictions and potentially generalize to an unseen number of objects. Although V-CDN is formulated to extract the causal structure of a fixed system through visual observations, we pretrain its perception module on our randomized \textit{YCBObjs} dataset for a direct comparison to our model. To perform a control task, we condition the V-CDN on the action by adding an encoding of the action vector to each keypoint embedding. Fig.~\ref{fig:ycb-mpc} compares the MPC results (normalized to the number of objects) of our method and V-CDN both trained on a subset of 3 YCB objects. The MPC performance of both models is comparable for rearranging 3 objects (green lines). However, V-CDN is most accurate for the number of objects it has been trained on and significantly less accurate when generalizing to an unseen number of objects. We attribute this to two reasons: (1) Although using keypoints, V-CDN is formulated to infer explicit causal structure and physical attributes for the graph representation of fixed environments which does not necessarily carry over to unseen circumstances. (2) Unlike our method, V-CDN does not take into account the visual features in the model and only uses the keypoint positions.


\begin{figure*}[h]
    \begin{center}
    \includegraphics[width= 1.00 \textwidth]{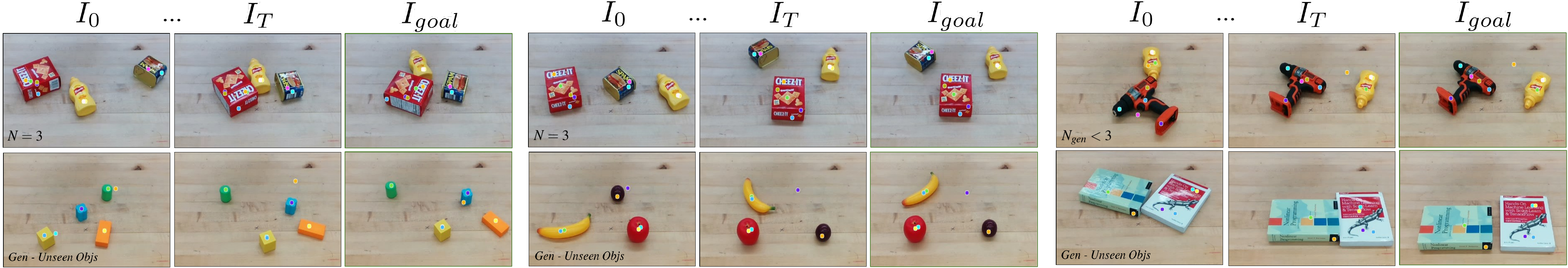}
    \end{center}
\vspace{-4 mm}
    \caption{Qualitative results in the real setting. Top row: Performance of the model on \textit{RealYCB} objects. Bottom row: Generalization to unseen objects.}
\label{fig:real-qual}
\vspace{-1 mm}
\end{figure*}

\begin{figure}[h]
\vspace{1 mm}
    \begin{center}
    \includegraphics[width= 0.95\linewidth]{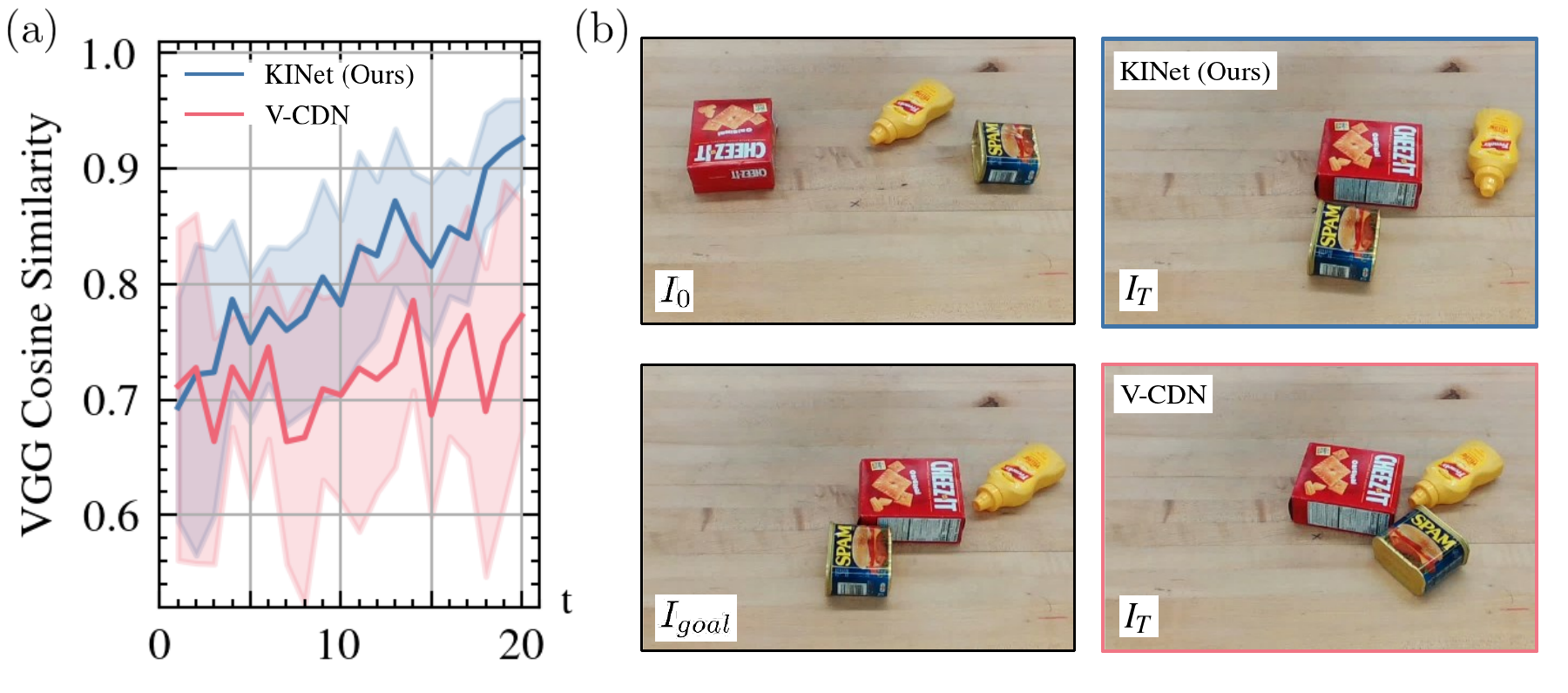}
\vspace{-6 mm}
    \end{center}
    \caption{Baseline comparison in the real setting. Our model more accurately rearranges the \textit{RealYCB} objects to match the goal image.}
\label{fig:real-base} 
\vspace{-1 mm}
\end{figure}

\subsection{Analysis and Ablation}
\label{sec:results:ablation}

We justify the major choices we made to formulate the model with ablation studies: the probabilistic graph representation (\textit{KINet - determ}), and the contrastive loss (\textit{KINet - no ctr}). As shown in Table~\ref{table:forw}, the best forward prediction for both \textit{TopView} and \textit{SideView} images is achieved when the model is probabilistic and trained with a contrastive loss. The contrastive loss is an essential element in our approach to ensure the learned forward model is accurately action conditional. Also, with a probabilistic graph representation, our model achieves better generalization compared to the deterministic variant. This performance gap is more evident when generalizing to unseen geometries (Table~\ref{table:gen}).

\ch{We ran additional experiments to examine the effect of the number of keypoints. Fig. \ref{fig:var-kps}-b shows the distance to the goal configuration after executing 30 planning steps with KINet, trained with different numbers of keypoints. Notably, for object rearrangement involving 3 YCB objects, KINet with fewer than 6 keypoints exhibited considerably larger errors. This indicates that in environments with complex visual features, optimal performance is achieved by a larger number of keypoints to ensure proper keypoint assignment (i.e., with $K=3$, Fig. \ref{fig:var-kps}-a shows one object remains unassigned).}

\ch{Since the number of objects is not hard-coded into our formulation, KINet allows for training on a variable number of objects. We experimented with training on varying numbers of \textit{YCBObjs} ($N$=1 to 5), as opposed to a fixed number ($N$=3). The results, shown in Figure \ref{fig:var-kps}-c, demonstrate that training KINet on a variable number of objects leads to a more balanced performance. As expected, for $N>3$, the final distance to the goal improved since the model was trained on these scenarios instead of generalizing. However, we did not observe a statistically significant change for $N\leq3$.}

\subsection{Real Robot Results} 

After training the model in \textit{YCBObjs} simulation, we test for the real robot performance (see Fig.~\ref{fig:setup}). Fig.~\ref{fig:real-qual} (top row) shows examples of the control task. The framework successfully transferred to the real setting and was able to perform the \textit{RealYCB} object rearrangement tasks. We observed that some of the detected keypoints were slightly less consistent compared to the simulation. We attribute this to the sim2real domain gap (e.g., the tabletop texture and camera noise).

We also tested for generalization to unseen objects (Fig.~\ref{fig:real-qual}, bottom row). Our model generalized to unseen objects by appropriately distributing the keypoints across the objects. We noticed that in some instances a keypoint is attached to the tabletop due to its visual feature. Regardless, the control task is achieved as the majority of keypoints are attached to the objects. The GraphMPC algorithm selects the action based on maximizing graph similarity to the goal scene across all nodes which reduces the effect of a single inconsistent keypoint on the rearrangement task.

We also compare with V-CDN baseline in the real setting. Since the ground-truth positions are unknown in the real world, we use image similarity to quantify the accuracy of the rearrangement task. Specifically, we used the pretrained VGG \cite{simonyan2014very} to measure the cosine similarity between the scene image and the goal image in the feature space to obtain MPC results for the real-world setting. As shown in Fig.~\ref{fig:real-base}, our model is consistently more accurate in rearranging the real scene and archives a significantly higher similarity to the goal image.


\section{Conclusion}
\vspace{-1 mm}
In this paper, we propose a method for learning action-conditioned forward models based only on RGB images. We showed that our approach effectively makes forward predictions with keypoint factorization of the scene image. Also, we showed that a keypoint-based forward model, unlike prior work, does not make assumptions about the number of objects which allows for automatic generalization to a variety of unseen circumstances. Importantly, our model is trained without any explicit supervision on ground-truth object states. 
\ch{Our framework has a limitation regarding the fixed number of expected keypoints. Although we demonstrate that this approach enhances generalizability compared to fixing the number of objects, determining the optimal number of keypoints for a specific environment requires experimenting with various numbers.
Furthermore, we observed slight inconsistencies in keypoints for real-robot. As a future direction, enhancing the keypoint extraction module to better handle real settings would be an interesting area of investigation.}


\bibliographystyle{plainnat}
\bibliography{references}


\end{document}